\documentclass[11pt,a4paper]{article}
\usepackage[hyperref]{acl2019}
\usepackage{times}
\usepackage{latexsym}
\usepackage{xspace}
\usepackage{url}
\usepackage{graphicx}

\newcommand{\glove}{\textsc{glove}\xspace}

\aclfinalcopy 

\title{Neural Legal Judgment Prediction in English}

\author{Ilias Chalkidis$\ast$  \qquad Ion Androutsopoulos$\ast$ \\ \textbf{Nikolaos Aletras$\ast\ast$} \\ $\ast$ Department of Informatics, Athens University of Economics and Business, Greece \\
$\ast\ast$ Computer Science Department, University of Sheffield, UK\\ 
{\tt {\normalsize[ihalk,ion]@aueb.gr, n.aletras@sheffield.ac.uk}}}

\date{}

\begin{document}
\maketitle
\begin{abstract}
Legal judgment prediction is the task of automatically predicting the outcome of a court case, given a text describing the case's facts. Previous work on using neural models for this task has focused on Chinese; only feature-based models (e.g., using bags of words and topics) have been considered in English. We release a new English legal judgment prediction dataset, containing cases from the European Court of Human Rights. We evaluate a broad variety of neural models on the new dataset, establishing strong baselines that surpass previous feature-based models in three tasks: (1) binary violation classification; (2) multi-label classification; (3) case importance prediction.
We also explore if models are biased towards demographic information via data anonymization.
As a side-product, we propose a hierarchical version of \textsc{bert}, which bypasses \textsc{bert}'s length limitation.
\end{abstract}

\section{Introduction}

Legal information is often represented in textual form (e.g., legal cases, contracts, bills).
Hence, legal text processing is a growing area in \textsc{nlp}  with various applications such as legal topic classification \cite{Nallapati2008,Chalkidis2019nllp}, court opinion generation \cite{Ye2018} and analysis \cite{Wang2012}, legal information extraction \cite{Chalkidis2018}, and entity recognition \cite{Cardellino2017,Chalkidis2017}. Here, we focus on \emph{legal judgment prediction}, where given a text describing the facts of a legal case, the goal is to predict the court's outcome \cite{Aletras2016,Sulea2017,Luo2017,Zhong2018,Hu2018}.

Such models may assist legal practitioners and citizens, while reducing legal costs and improving access to justice~\cite{Lawlor1963,Katz2012,Stevenson2015}. Lawyers and judges can use them to estimate the likelihood of winning a case and come to more consistent and informed judgments, respectively. Human rights organizations and legal scholars can employ them to scrutinize the fairness of judicial decisions unveiling if they correlate with biases~\cite{Doshi2017,Binns2018}.

This paper contributes a new publicly available English legal judgment prediction dataset of cases from the European Court of Human Rights (\textsc{echr}).\footnote{The dataset is submitted at \url{https://archive.org/details/ECHR-ACL2019}.} Unlike \citet{Aletras2016}, who provide only features from approx.\ 600 \textsc{echr} cases, our dataset is substantially larger ($\sim$11.5k cases) and provides access to the raw text. As a second contribution, we evaluate several neural models in legal judgment prediction for the first time in English. We consider three tasks: (1) binary classification (i.e., violation of a human rights article or not), the only task considered by \citet{Aletras2016}; (2) multi-label classification (type of violation, if any); (3) case importance detection. In all tasks, neural models outperform an \textsc{svm} with bag-of-words \citep{Aletras2016,Medvedeva2018}, the only method tested in English legal judgment prediction so far. As a third contribution, we use an approach based on data anonymization to study, for the first time, whether the legal predictive models are biased towards demographic information or factual information relevant to human rights. Finally, as a side-product, we propose a hierarchical version of \textsc{bert} \cite{BERT}, which bypasses \textsc{bert}'s length limitation 
and leads to the best results.

\section{ECHR Dataset}

\textsc{echr} hears allegations that a state has breached human rights provisions of the European Convention of Human Rights.\footnote{An up-to-date copy of the European Convention of Human Rights is available at \url{https://www.echr.coe.int/Documents/Convention_ENG.pdf}.} Our dataset contains 
approx.\ 11.5k cases from \textsc{echr}'s public database.\footnote{See \url{https://hudoc.echr.coe.int}. Licensing conditions are compatible with the release of our dataset.} For each case, the dataset provides a list of \emph{facts} extracted using regular expressions from the case description, as in \citet{Aletras2016}\footnote{Using regular expressions to segment legal text from \textsc{echr}
is usually trivial, as the text has a specific structure. See an example from \textsc{echr}'s Data Repository (\url{http://hudoc.echr.coe.int/eng?i=001-193071}).} (see Fig.~\ref{fig:attention}). Each case is also mapped to \emph{articles} of the Convention that were violated (if any). An \emph{importance score} is also assigned by \textsc{echr} (see Section~\ref{sec:tasks}). 

The dataset is split into training, development, and test sets (Table \ref{tab:dataset}). The training and development sets contain cases from 1959 through 2013, and the test set from 2014 through 2018. The training and development sets are balanced, i.e., they contain equal numbers of cases with and without violations. We opted to use a balanced training set to make sure that our data and consequently our models are not biased towards a particular class. The test set contains more (66\%) cases with violations, which is the approximate ratio of cases with violations in the database. We also note that 45 out of 66 labels are not present in the training set, while another 11 are present in fewer than 50 cases. Hence, the dataset of this paper is also a good testbed for few-shot learning.

\section{Legal Prediction Tasks}
\label{sec:tasks}

\subsection{Binary Violation} 
Given the facts of a case, we aim to classify it as positive if \emph{any} human rights article or protocol has been violated and negative otherwise.

\subsection{Multi-label Violation} 
Similarly, the second task is to predict which specific human rights articles and/or protocols have been violated (if any). The total number of articles and protocols of the European Convention of Human Rights are 66 up to day. For that purpose, we define a multi-label classification task where no labels are assigned when there is no violation.

\subsection{Case Importance}
We also predict the importance of a case on a scale from 1 (key case) to 4 (unimportant) in a regression task. These scores, provided by the \textsc{echr}, denote a case's contribution in the development of case-law allowing legal practitioners to identify pivotal cases. Overall in the dataset, the scores are: 1 (1096 documents), 2 (904), 3 (2,982) and 4 (6,496), indicating that approx. 10\% are landmark cases, while the vast majority (83\%) are considered more or less unimportant for further review.

\begin{table}
\centering
\footnotesize
\begin{tabular}{lcccc}
\hline
  Subset & Cases ($C$) & Words/$C$ & Facts/$C$ & Articles/$C$ \\
\hline
  Train & 7,100 & 2,421 & 43 & 0.71\\
  Dev. & 1,380 & 1,931 & 30 & 0.96 \\
  Test & 2,998 & 2,588 & 45 & 0.71\\
  \hline
\end{tabular}
\caption{Statistics of the \textsc{echr} dataset. The size of the label set (ECHR articles) per case ($C$) is $L=66$.}
\vspace*{-4mm}
\label{tab:dataset}
\end{table}

\section{Neural Models}

\paragraph{BiGRU-Att:}
The fisrt model is a \textsc{bigru} with self-attention \citep{Xu2015} where the facts of a case are concatenated into a word sequence. Words are mapped to embeddings and passed through a stack of \textsc{bigru}s. A single case embedding ($h$) is computed as the sum of the resulting context-aware embeddings ($\sum_i a_i h_i$) weighted by self-attention scores ($a_i$). The case embedding ($h$) is passed to the output layer using a sigmoid for binary violation, softmax for multi-label violation, or no activation for case importance regression.

\paragraph{HAN:}
The Hierarchical Attention Network \citep{Yang2016} is a state-of-the-art model for text classification. We use a slightly modified version where a \textsc{bigru} with self-attention reads the words of each fact, as in \textsc{bigru-att}, producing fact embeddings. A second-level \textsc{bigru} with self-attention reads the fact embeddings, producing a single case embedding that goes through a similar output layer as in \textsc{bigru-att}. 

\paragraph{LWAN:}
The Label-Wise Attention Network \citep{Mullenbach2018} has been shown to be robust in multi-label classification. Instead of a single attention mechanism, \textsc{lwan} employs $L$ attentions, one for each possible label. This produces $L$ case embeddings ($h^{(l)} = \sum_i a_{l,i} h_i$) 
per case, each one specialized to predict the corresponding label. Each of the case embeddings goes through a separate linear layer ($L$ linear layers in total), each with a sigmoid, to decide if the corresponding label should be assigned. Since this is a multi-label model, we use it only in multi-label violation. 

\paragraph{BERT and HIER-BERT:}
\textsc{bert} \citep{BERT} is a language model based on Transformers \citep{Vaswani2017} pretrained on large corpora. For a new task, a task-specific layer is added on top of \textsc{bert} and is trained jointly by fine-tuning on task-specific data. We add a linear layer on top of \textsc{bert}, with a sigmoid, softmax, or no activation, for binary violation, multi-label violation, and case importance, respectively.\footnote{The extra linear layer is fed with the `classification' token of the \textsc{bert-base} version of \citet{BERT}.} \textsc{bert} can process texts up to 512 wordpieces, whereas our case descriptions are up to 2.6k words, thus we truncate them to \textsc{bert}'s maximum length, which affects its performance. This also highlights an important limitation of \textsc{bert} in processing long documents, a common characteristic in legal text processing.

To surpass \textsc{bert}'s maximum length limitation, we also propose a hierarchical version of \textsc{bert} 
(\textsc{hier-bert}). Firstly \textsc{bert-base} reads the words of each fact, producing fact embeddings. Then a self-attention mechanism reads fact embeddings, producing a single case embedding that goes through a similar output layer as in \textsc{han}.

\section{Experiments}

\subsection{Experimental Setup}

\paragraph{Hyper-parameters:} We use pre-trained
\glove \citep{pennington2014glove} embeddings ($d=200$) for all experiments. Hyper-parameters are tuned by random sampling 50 combinations and selecting the values with the best development loss in each task.\footnote{Ranges: \textsc{GRU} hidden units \{200, 300, 400\}, number of stacked \textsc{bigru} layers \{1, 2\}, batch size \{8, 12, 16\}, dropout rate \{0.1, 0.2, 0.3, 0.4\}, word dropout rate \{0.0, 0.01, 0.02\}.} Given the best hyper-parameters, we perform five runs for each model reporting mean scores and standard deviations. We use categorical cross-entropy loss for the classification tasks and mean absolute error for the regression task, Glorot initialization \cite{Glorot2010}, Adam \cite{Kingma2015} with default learning rate 0.001, and early stopping on the development loss.  

\paragraph{Baselines:} 
A majority-class (\textsc{majority}) classifier is used in binary violation and case importance. A second baseline (\textsc{coin-toss}) randomly predicts violation or not in binary violation task. 
We also compare our methods against a linear \textsc{svm} with bag-of-words features (most frequent [1, 5]-grams across all training cases weighted by \textsc{tf-idf}), dubbed \textsc{bow-svm}, similar to \citet{Aletras2016} and \citet{Medvedeva2018} for the binary task; multiple one-vs-rest classifiers for the multi-label task; and Support Vector Regression (\textsc{bow-svr}) for the case importance prediction.\footnote{We tune the hyper-parameters of \textsc{bow-svm/svr} and select kernel (\textsc{rbf}, linear) with a grid search on the dev.\ set.}

\begin{table}[t!]
\centering
\footnotesize
\begin{tabular}{lccc}
  \hline
  & {P} & {R} & {F1}  \\
  \hline
   \textsc{majority} & 32.9 $\pm$ 0.0 & 50.0 $\pm$ 0.0 & 39.7 $\pm$ 0.0 \\
   \textsc{coin-toss} & 50.4 $\pm$ 0.7 & 50.5 $\pm$ 0.8 & 49.1 $\pm$ 0.7 \\
   \hline
   \multicolumn{4}{c}{\bf Non-Anonymized}\\
  \hline
  \textsc{bow-svm} & 71.5 $\pm$ 0.0 & 72.0 $\pm$ 0.0 & 71.8  $\pm$ 0.0\\
  \textsc{bigru-att} & 87.1 $\pm$ 1.0 & 77.2 $\pm$ 3.4 & 79.5 $\pm$ 2.7\\
  \textsc{han} & 88.2 $\pm$ 0.4  & 78.0  $\pm$ 0.2 & 80.5 $\pm$ 0.2\\
  \textsc{bert} & 24.0 $\pm$ 0.2 & 50.0 $\pm$ 0.0 & 17.0 $\pm$ 0.5\\
  \textsc{hier-bert}  & \textbf{90.4}  $\pm$ 0.3  & \textbf{79.3}  $\pm$ 0.9 & \textbf{82.0} $\pm$ 0.9  \\
  \hline
  \multicolumn{4}{c}{\bf Anonymized}\\
  \hline
  \textsc{bow-svm} & 71.6 $\pm$ 0.0  & 70.5 $\pm$ 0.0 & 70.9 $\pm$ 0.0\\
  \textsc{bigru-att} & \textbf{87.0} $\pm$ 1.0 & 76.6 $\pm$ 1.9 & 78.9 $\pm$ 1.9\\
  \textsc{han} & 85.2 $\pm$ 4.9 & \textbf{78.3} $\pm$ 2.0 & \textbf{80.2} $\pm$ 2.7\\
  \textsc{bert} & 17.0 $\pm$ 3.0 & 50.0 $\pm$ 0.0 & 25.4 $\pm$ 0.4\\
  \textsc{hier-bert}  & 85.2  $\pm$ 0.3  & 78.1  $\pm$ 1.3 & 80.1 $\pm$ 1.1  \\
  \hline
\end{tabular}
\caption{Macro precision (P), recall (R), F1 for the \textbf{binary violation} prediction task ($\pm$ std. dev.).}
\vspace*{-5mm}
\label{tab:binaryresults}
\end{table}

\begin{figure*}[!t]
  \centering
    \includegraphics[scale=0.52]{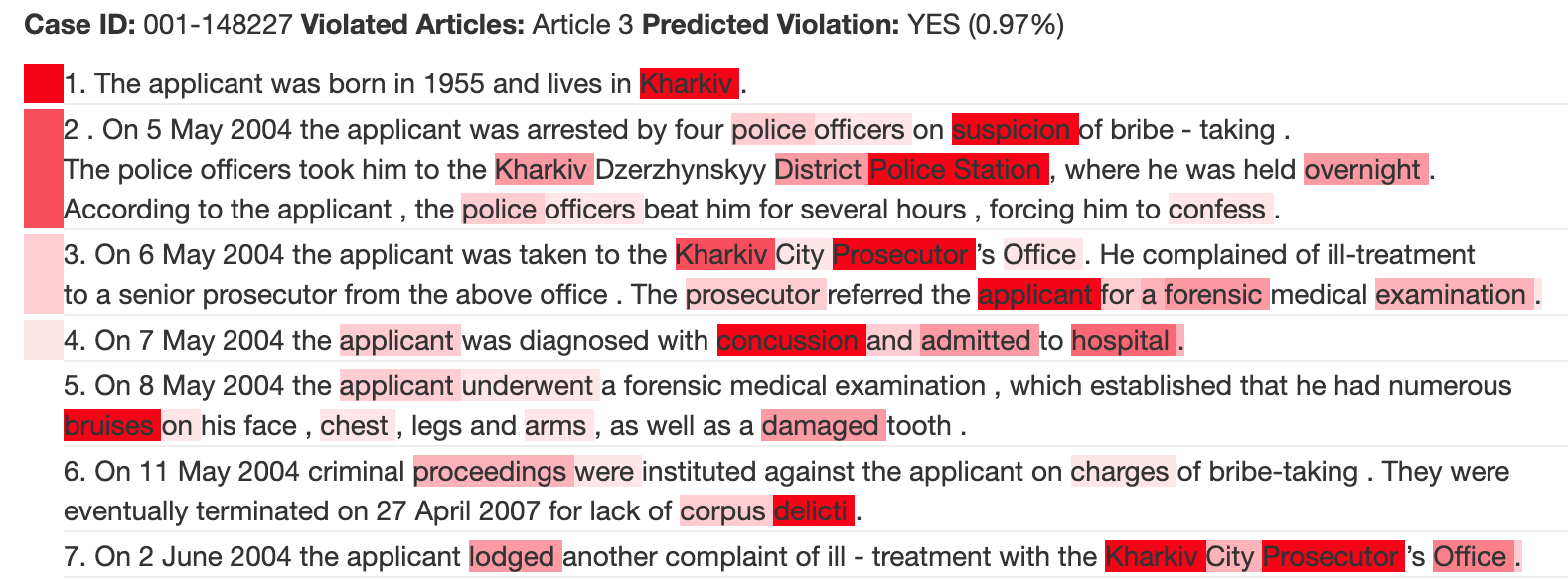}
  \caption{Attention over words (colored words) and 
  facts (vertical heat bars) as produced by \textsc{han}.}
  \vspace*{-4mm}
  \label{fig:attention}
\end{figure*}

\subsection{Binary Violation Results}

Table~\ref{tab:binaryresults} (upper part) shows the results for binary violation. We evaluate models using macro-averaged precision (P), recall (P), F1. The weak baselines (\textsc{majority}, \textsc{coin-toss}) are widely outperformed by the rest of the methods. \textsc{bigru-att} outperforms in 
F1 (79.5 vs.\ 71.8) the previous best performing method  \citep{Aletras2016} in English judicial prediction. This is aligned with results in Chinese \cite{Luo2017,Zhong2018,Hu2018}. \textsc{han} slightly improves over \textsc{bigru-att} (80.5 vs.\ 79.5), while being more robust across runs (0.2\% vs.\ 2.7\% std. dev.). \textsc{bert}'s poor performance is due to the truncation of case descriptions, while \textsc{hier-bert} that uses the full case leads to the best results. We omit \textsc{bert} from the following tables, since it performs poorly.

Fig.~\ref{fig:attention} shows the attention scores over words and facts of 
\textsc{han} for a case that \textsc{echr} found to violate Article 3, which prohibits torture and `inhuman or degrading treatment or punishment'. 
Although fact-level attention wrongly assigns high attention to the first fact, which seems irrelevant, it then successfully focuses on facts 2--4, which report that police officers beat the applicant for several hours, that the applicant complained, was referred for forensic examination, diagnosed with concussion etc. Word attention also successfully focuses on words like `concussion', `bruises', `damaged', but it also highlights entities like `Kharkiv', its `District Police Station' and `City Prosecutor's office', which may be indications of bias.

\paragraph{Models Biases:}
We next investigate how sensitive our models are to demographic information appearing in the facts of a case. Our assumption is that an unbiased model should not rely on information about nationality, gender, age, etc. To test the sensitivity of our models to such information, we train and evaluate them in an anonymized version of the dataset. The data is anonymized by using \textsc{spacy}'s (\url{https://spacy.io}) Named Enity Recognizer, replacing all recognized entities with type tags (e.g., `Kharkiv' $\rightarrow$ \textsc{location}). 

While neural methods seem to exploit named entities among other information, as in Figure~\ref{fig:attention}, the results in Table~\ref{tab:binaryresults} indicate that performance is comparable even when this information is masked, with the exception of \textsc{hier-bert} that has quite worse results (2\%) compared to using non-anonymized data, suggesting model bias. We speculate that \textsc{hier-bert} is more prone to over-fitting compared to the other neural methods that rely on frozen \glove embeddings, because the embeddings of \textsc{bert}'s wordpieces are trainable and thus can freely adjust to the vocabulary of the training documents including demographic information. 

\begin{table}[t!]
\centering
\footnotesize
\begin{tabular}{lccc}
  \hline
  \multicolumn{4}{c}{\textsc{overall} (all labels)} \\ 
  \hline
  & {\bf P} & {\bf R} & {\bf F1} \\
  \hline
  \textsc{bow-svm} & 56.3  $\pm$ 0.0 & 45.5 $\pm$ 0.0 & 50.4 $\pm$ 0.0  \\
  \textsc{bigru-att} & 62.6  $\pm$ 1.2 & 50.9  $\pm$ 1.5 & 56.2 $\pm$ 1.3 \\
  \textsc{han} & 65.0  $\pm$ 0.4 & \textbf{55.5}  $\pm$ 0.7 & 59.9  $\pm$ 0.5 \\
  \textsc{lwan}  & 62.5  $\pm$ 1.0 & 53.5  $\pm$ 1.1 & 57.6 $\pm$ 1.0  \\
  \textsc{hier-bert}  & \textbf{65.9} $\pm$ 1.4 & 55.1 $\pm$ 3.2 & \textbf{60.0} $\pm$ 1.3 \\
  \hline
  \multicolumn{4}{c}{\textsc{frequent} ($\geq$50)} \\ 
  \hline
  \textsc{bow-svm} & 56.3 $\pm$ 0.0 & 45.6 $\pm$ 0.0 & 50.4 $\pm$ 0.0 \\
  \textsc{bigru-att} & 62.7  $\pm$ 1.2 & 52.2  $\pm$ 1.6 & 57.0  $\pm$ 1.4 \\
  \textsc{han} & 65.1  $\pm$ 0.3 & \textbf{57.0}  $\pm$ 0.8 & \textbf{60.8}  $\pm$ 1.3 \\
  \textsc{lwan} & 62.8  $\pm$ 1.2 & 54.7  $\pm$ 1.2 & 58.5 $\pm$ 1.0  \\
   \textsc{hier-bert}  & \textbf{66.0} $\pm$ 1.4 & 56.5 $\pm$ 3.3 & \textbf{60.8} $\pm$ 1.3 \\
  \hline
  \multicolumn{4}{c}{\textsc{few} ([1,50))} \\ 
   \hline
  \textsc{bow-svm}   & - & - & - \\
  \textsc{bigru-att} & 36.3  $\pm$ 13.8 & 03.2 $\pm$ 23.1 & 05.6  $\pm$ 03.8\\
  \textsc{han} & 30.2 $\pm$ 35.1 & 01.6  $\pm$ 01.2 & 02.8 $\pm$ 01.9\\
  \textsc{lwan} & 24.9 $\pm$ 06.3 & \textbf{07.0} $\pm$ 04.1 & \textbf{10.6} $\pm$ 05.2 \\
   \textsc{hier-bert}  & \textbf{43.6} $\pm$ 14.5 & 05.0 $\pm$ 02.8 & 08.9 $\pm$ 04.9 \\
  \hline
\end{tabular}
\caption{Micro precision, recall, F1 in \textbf{multi-label violation} for all, frequent, and few training instances.}
\vspace*{-5mm}
\label{tab:multilabel}
\end{table}
\subsection{Multi-label Violation Results}

Table~\ref{tab:multilabel} reports micro-averaged precision (P), recall (R), and F1 results for all methods, now including \textsc{lwan}, in multi-label violation prediction. The results are also grouped by label frequency for all (\textsc{overall}), \textsc{frequent}, and \textsc{few} labels (articles), counting frequencies on the training subset. 

We observe that predicting specific articles that have been violated is a much more difficult task than predicting if \emph{any} article has been violated in a binary setup (cf.\ Table~\ref{tab:binaryresults}). Overall, \textsc{hier-bert} outperforms \textsc{bigru-att} and \textsc{lwan} (60.0 vs.\ 57.6 micro-F1), which is tailored for multi-labeling tasks, while being comparable with \textsc{han} (60.0 vs.\ 59.9 micro-F1).
All models under-perform in labels with \textsc{few} training examples, demonstrating the difficulty of few-shot learning in \textsc{echr} legal judgment prediction. The main reason is that labels in the \textsc{few} group, 11 in total, are extremely rare and have been assigned in 1.25\% of the documents across all datasets, while the most frequent 4 labels overall (Articles 3, 5, 6 and 13) have been assigned in approx.\ 42\% of the documents.

\subsection{Case Importance Results}

Table~\ref{tab:importance} shows the mean absolute error (\textsc{mae}) obtained when predicting case importance. Surprisingly, \textsc{majority} outperforms the rest of the methods. As already noted in Section \ref{sec:tasks}, the distribution of importance scores is highly skewed in favour of the majority class, thus \textsc{majority} can correctly predict the score in most cases with zero mean absolute error (\textsc{mae}). \textsc{bow-svr} performs worse than \textsc{bigru-att}, while \textsc{han} is 10\% and 3\% better, respectively. \textsc{hier-bert} further improves the results, outperforming \textsc{han} by 17\%. 

While \textsc{majority} has the lowest mean absolute error, it cannot distinguish important from unimportant cases, thus it is practically useless. To evaluate the methods on that matter, we measure the correlation between the gold scores and each method's predictions with \textsc{spearman}'s $\rho$. \textsc{hier-bert} has the best $\rho$ (.527), 
indicating a moderate positive correlation ($>0.5$), which is not the case for the rest of the methods. The overall results indicate that a case's importance cannot be predicted solely by the case facts and possibly also relies on background knowledge (e.g., judges' experience, court's history, rarity of article's violation).

\begin{table}[!t]
\centering
\small
\begin{tabular}{lcc}
  \hline
   & \textsc{MAE} & \textsc{spearman}'s $\rho$ \\ 
  \hline
  \textsc{majority} & \textbf{.369} $\pm$ .000 & $N/A$* \\
  \textsc{bow-svr}  & .585  $\pm$ .000  & .370 $\pm$ .000 \\
  \textsc{bigru-att}  & .539  $\pm$ .073  & .459 $\pm$ .034 \\
  \textsc{han}  & .524  $\pm$ .049  & .437 $\pm$ .018  \\
   \textsc{hier-bert}  & .437 $\pm$ .018  & \textbf{.527} $\pm$ .024\\
  \hline
\end{tabular}
\caption{Mean Absolute Error and Spearman's $\rho$ for {\bf case importance}. Importance ranges from 1 (most important) to 4 (least). * Not Applicable.}
\vspace*{-4mm}
\label{tab:importance}
\end{table}

\subsection{Discussion}

We can only speculate that \textsc{han}'s fact embeddings distill importance-related features from each fact, allowing its second-level \textsc{gru} to operate on a sequence of fact embeddings that are being exploited by the fact-level attention mechanism and provide a more concise view of the entire case. The same applies to \textsc{hier-bert}, which relies on \textsc{bert}'s fact embeddings and the same fact-level attention mechanism.
By contrast, \textsc{bigru-att} operates on a single long sequence of concatenated facts, making it more difficult for its \textsc{bigru} to combine information from multiple, especially distant, facts. This may explain the good performance of \textsc{han} and \textsc{hier-bert} across all tasks.

\section{Related Work}
Previous work on legal judgment prediction in English used linear models with features based on bags of words and topics to represent legal textual information extracted from cases \citep{Aletras2016,Medvedeva2018}. 

More sophisticated neural models have been considered only in Chinese. \citet{Luo2017} use \textsc{han}s to encode the facts of a case and a subset of predicted relevant law articles to predict criminal charges that have been manually annotated. In their experiments, the importance of few-shot learning is not taken into account since the criminal charges that appear fewer than 80 times are filtered out. However in reality, a court is able to judge even under rare conditions. \citet{Hu2018} focused on few-shot charges prediction using a multi-task learning scenario, predicting in parallel a set of discriminative attributes as an auxiliary task. Both the selection and annotation of these attributes are manually crafted and dependent to the court. \citet{Zhong2018} decompose the problem of charge prediction into different subtasks that are tailored to the Chinese criminal court using multi-task learning. 

\section{Limitations and Future Work}
The neural models we considered outperform previous feature-based models, but provide no justification for their predictions. Attention scores (Fig.~\ref{fig:attention}) provide some indications of which parts of the texts affect the predictions most, but are far from being justifications that legal practitioners could trust; see also \citet{Jain2019}. Providing valid justifications is an important priority for future work and an emerging topic in the NLP community.\footnote{\url{http://aclweb.org/anthology/W18-5400}} In this direction, we plan to expand the scope of this study by exploring the automated analysis of additional resources (e.g., relevant case law, dockets, prior judgments) that could be then utilized in a multi-input fashion to further improve performance and justify system decisions. We also plan to apply neural methods to data from other courts, e.g., the European Court of Justice, the US Supreme Court, and multiple languages, to gain a broader perspective of their potential in legal justice prediction. Finally, we plan to adapt bespoke models proposed for the Chinese Criminal Court \citep{Luo2017,Zhong2018,Hu2018} to data from other courts and explore multi-task learning.

\section*{Acknowledgements}
This work was partly supported by the Research Center of the Athens University of Economics and Business.

\bibliography{acl2019}
\bibliographystyle{acl_natbib}

\end{document}